\pdfoutput=1

\documentclass[11pt]{article}

\usepackage[]{acl}

\usepackage{times}
\usepackage{latexsym}
\usepackage{amsmath}
\usepackage[T1]{fontenc}

\usepackage[utf8]{inputenc}

\usepackage{microtype}

\usepackage{multirow} 
\usepackage{soul}
\usepackage{float}
\usepackage{subfigure}
\usepackage{tabularx}
\usepackage{graphicx} 
\usepackage{booktabs}
\usepackage{algpseudocode}
\usepackage{listings}
\newfloat{Code}{htbp}{Code}
\usepackage{caption}
\usepackage{xcolor}
\usepackage{algorithm}
\usepackage[colorinlistoftodos,prependcaption,textsize=tiny]{todonotes}
\usepackage{xspace}
\usepackage{xargs} 
\newcommandx{\synset}[1]{{\fontfamily{lmtt}\selectfont \textbf{#1}}}
\newcommandx{\alex}[2][1=]{\todo[linecolor=red,backgroundcolor=red!20,bordercolor=red,#1,size=tiny]{AS: #2}\xspace}
\title{Better Language Model with Hypernym Class Prediction}

\author{He Bai\Thanks{Most of the work was done during the internship at Microsoft Research. Code: \url{https://github.com/richardbaihe/robustLM.git}} \\
  University of Waterloo \\
  \texttt{he.bai@uwaterloo.ca} \\\And
  Tong Wang \\
  Microsoft Research\\
  \texttt{Tong.Wang@microsoft.com} \\
  \AND
  Alessandro Sordoni \\
  Microsoft Research \\
  \texttt{alsordon@microsoft.com} \\\And
  Peng Shi\\
  University of Waterloo \\
  \texttt{peng.shi@uwaterloo.ca} \\}

\newcommand{\arxiv}{\textbf{\textsc{arXiv}}}
\newcommand{\wikitext}{\textbf{WikiText-103}}
\newcommand{\blahfull}{Hypernym Class Prediction\xspace}
\newcommand{\blah}{HCP\xspace}
\begin{document}
\maketitle
\begin{abstract}
Class-based language models (LMs) have been long devised to address context sparsity in $n$-gram LMs. 
In this study, we revisit this approach in the context of neural LMs. 
We hypothesize that class-based prediction leads to an implicit context aggregation for similar words and thus can improve generalization for rare words. 
We map words that have a common WordNet hypernym to the same class and train large neural LMs by gradually annealing from predicting the class to token prediction during training. 
Empirically, this curriculum learning strategy consistently improves perplexity over various large, highly-performant state-of-the-art Transformer-based models on two datasets,~\wikitext~and~\arxiv. 
Our analysis shows that the performance improvement is achieved without sacrificing performance on rare words. 
Finally, we document other attempts that failed to yield empirical gains, and discuss future directions for the adoption of class-based LMs on a larger scale.
\end{abstract}

\section{Introduction}
Over the course of the past decades, language modeling (LM) has transitioned from $n$-gram to neural models \citep{bengio2003neural,mnih2007three,devlin2018bert,brown2020language}.
Performance improvement of today's neural LMs is often achieved at the cost of increased computational resources.
For example, to capture long-term dependencies, various extensions of Transformer-based LMs have been proposed~\citep{DBLP:conf/acl/DaiYYCLS19,DBLP:conf/iclr/RaePJHL20}.
These modifications bring about significant improvements on held-out perplexity, but training cost also increases significantly due to large GPU memory consumption and more computations at each training step.

In parallel, alternative training strategies have also been proposed~\cite{guu2020realm,ziegler2019latent,residual20}. In this paper, we explore the effectiveness of class-based language models~(CLMs, \citealt{brown1992class}) in the context of neural LMs. CLMs group individual words into coarser-grained classes and has proven effective in alleviating context sparsity in $n$-gram LMs~\citep{dagan1999similarity}. It has been also used to improve computational efficiency in neural LMs~\citep{morin2005hierarchical, pmlr-v70-grave17a}. More recently,~\citet{levine2019sensebert} pretrain masked LMs~\cite{devlin2018bert} by predicting WordNet supersense labels. However, the work focuses on word-sense disambiguation tasks and doesn't provide clear evidence of gains in terms of perplexity.

\begin{figure}[t]
 \centering
 \includegraphics[width=0.5\textwidth]{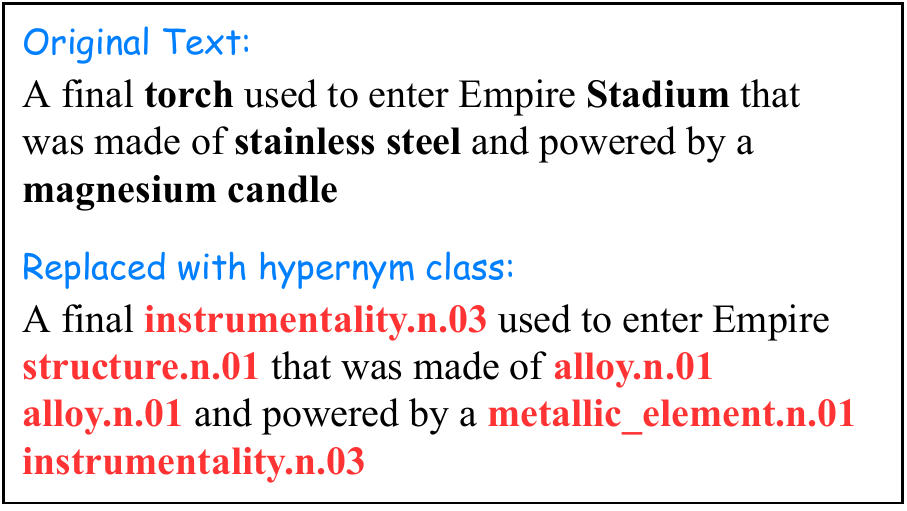}
 \caption{An example of word prediction training text and hypernym class prediction training text.}
 \label{fig:instruction}
\end{figure}

In this paper, we revisit CLM and assign words to classes by leveraging hypernym relations from the WordNet~\citep{miller1995wordnet}. Our proposal, dubbed \blahfull (\blah) is simple and effective: for each batch, we substitute a subset of the tokens with their WordNet hypernyms (see Figure~\ref{fig:instruction}). Then, we train an autoregressive LM on the resulting sentences using a mixed vocabulary composed of hypernyms and tokens. Crucially, we anneal the substitution rate during training,~i.e., we gently switch from hypernym prediction to token prediction, following a curriculum learning approach. Note that this approach does not require WordNet information at inference time nor increases training time.

Our approach is motivated by two hypotheses.
Firstly, mapping words to their hypernyms gives rise to a natural gradation of difficulty in the prediction task.
Prior work has shown that LM benefits from training on instances of increasing difficulty~\citep{bengio2009curriculum,press2020shortformer}.
We thus postulate that, when coupled with the right curriculum, \blah can improve LM training and perplexity. Secondly, we hypothesize that \blah can improve rare word generalization through implicit context sharing. 
Neural models still struggle to learn reliable representations for rare words~\cite{schick2019rare}.
With CLM-based models, data sparsity for rare words can be abated, e.g., when the representation of their contexts are potentially drawn closer to those of their more frequent siblings by way of label (hypernym) sharing.

Empirically, the proposed method consistently yields about 0.6--1.9\% relative reduction in perplexity over baselines on the~\wikitext~dataset~\cite{merity2016pointer}, and 1.3--3.1\% on the~\arxiv~dataset~\cite{lazaridou2021pitfalls}. These improvements are observed with respect to memory-augmented~\cite{DBLP:conf/acl/DaiYYCLS19} and segment-aware~\cite{Bai_Shi_Lin_Xie_Tan_Xiong_Gao_Li_2021} LMs. Importantly, the proposed method improves performance for both rare and frequent words. We also observe that this is in contrast with performance improvements in regular LMs, which seem to be achieved at the cost of worsened performance on rare words.

To the best of our knowledge, this is the first work that shows how perplexity of Transformer LMs can be improved by leveraging hypernymy relationships. We provide an extensive ablation study highlighting crucial elements of \blah. Amongst those, we found particularly important to adopt a curriculum learning approach, rather than multi-objective learning or adaptive-softmax, and excluding frequent words from the hypernym prediction task. We highlight the simplicity and effectiveness of the proposed method as our main contribution, and hope this study would facilitate further exploration in this line of research.





\section{Related Work}
Transformer-based models are now popular language models.
\citet{DBLP:conf/acl/DaiYYCLS19} propose Transformer-XL by extending the vanilla Transformer~\cite{vaswani2017attention} with a memory segment, which can encode more context tokens to predict the next token.
\citet{DBLP:conf/iclr/RaePJHL20} extend Transformer-XL with a compressed memory segment to further encode long-time context memory. Other works explore different sparse Transformers to encode much longer sequences for LM~\cite{beltagy2020longformer,roy2021routing}.
\citet{Bai_Shi_Lin_Xie_Tan_Xiong_Gao_Li_2021} propose a segment-aware Transformer~(Segatron) to encode more positional information for language modeling. Despite their effectiveness, neural models still struggle to learn reliable representations for rare words.
Some approaches have been proposed to tackle this challenge by way of morphology~\citep{luong2013better}, lexical similarity~\citep{rare_word_enriching_2019}, context similarity~\citep{schick2019rare,Khandelwal2020Generalization} and tokenization~\citep{kudo2018sentencepiece}.

In addition to the model modifications, other work investigated 
curriculum learning to train LMs. \citet{bengio2009curriculum} first find that curriculum learning could benefit LM training by training with high-frequency tokens first and low-frequency tokens later.
\citet{wu2020curricula} find that curricula works well when the training data is noisy or the training data is too large to iterate multiple epochs. 
\citet{press2020shortformer} find that training Transformer-based LMs with short sequences first could improve convergence speed and perplexity.

Related work aimed at integrating WordNet information into pretrained language models.~\citet{levine2019sensebert} propose SenseBERT by adding the word sense~(WordNet supersense) prediction as an additional task during BERT~\cite{devlin2018bert} pre-training. SenseBERT outperforms BERT on both word supersense disambiguation~\cite{raganato-etal-2017-wsd} task and word in context~\cite{pilehvar-camacho-collados-2019-wic} task. Recently,~\citet{porada2021modeling} use WordNet hypernymy chains as input to a Roberta~\cite{liu2019roberta} model to predict the plausibility of input events. In this work, our focus is to improve performance of auto-regressive LMs. We show that a multi-task strategy harms performance in this setting, and give a successful recipe to consistently boost LM performance with class-based predictions.

\begin{figure*}[ht!]
  \centering
  \includegraphics[height=0.35\textwidth ]{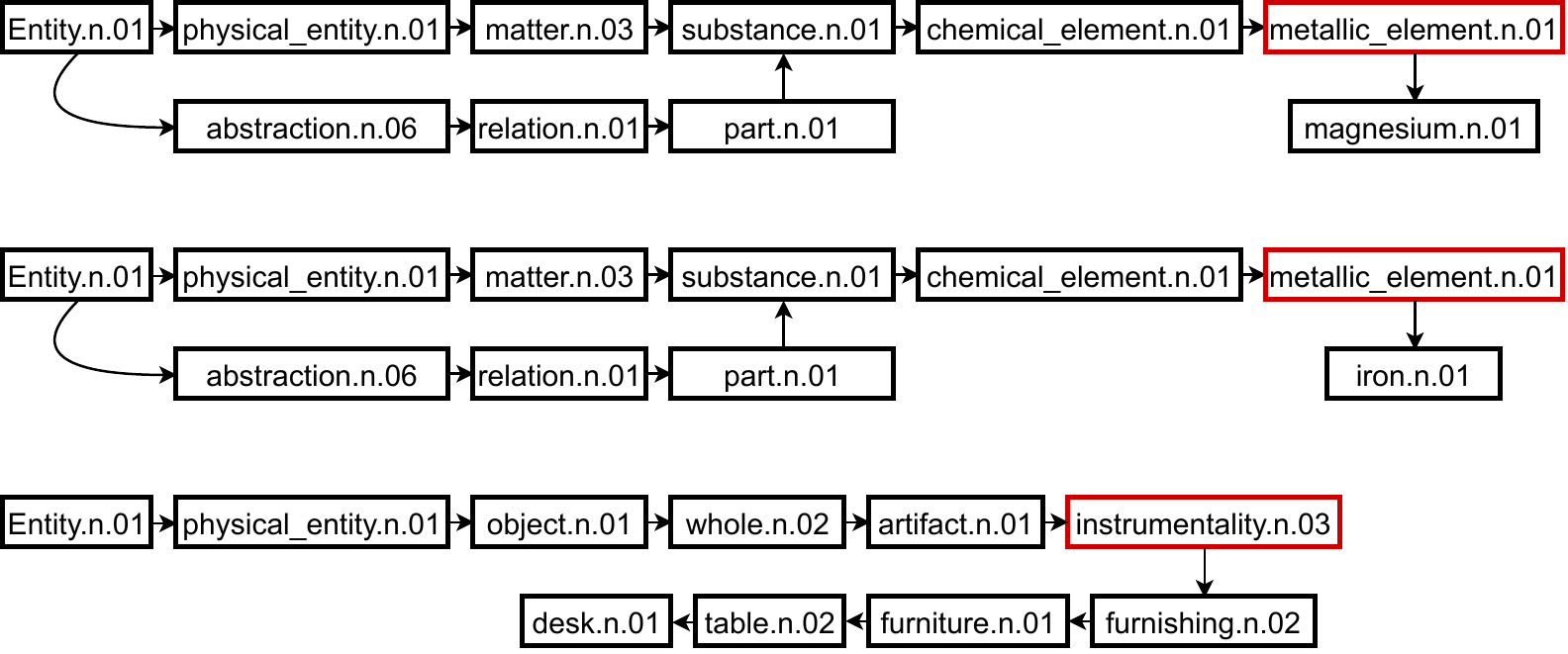}
  \caption{Hypernym-paths of synsets ``magnesium.n.01'', ``iron.n.01'', and ``desk.n.01'', corresponding to the word \emph{magnesium}, \emph{iron}, and \emph{desk} respectively.}
  \label{fig: synset}
\end{figure*}

\begin{figure}[ht!]
\begin{lstlisting}[mathescape,language=Python]
def token2class(token2freq, d, f):
    # token2freq is a dictionary whose key is the token and value is the tokens' occurrences)
    # d is the depth, f is the occurrence threthold
    rtn = {}
    for token, freq in token2freq.items():
        if freq > f:
            continue
        for synset in wordnet.synsets(token):
            for path in synset.hypernym_paths():
                if len(path)>=d and 'noun' in path[d-1]:
                    rtn[token] = path[d-1]
                    break
            if token in rtn:
                break
    return rtn
\end{lstlisting}
\captionof{Code}{Pseudocode for token to class mapping.}
\label{code: token2class}
\end{figure}

\section{Method}
Coupling class-based LM (CLM) and curriculum learning, \blah is to gradually anneal class prediction to token prediction during LM training.
In this section, we first describe how we instantiate word classes by leveraging hypernym relation from the WordNet.
We then present how to incorporate the proposed \blahfull task into LM training via curriculum learning.

\subsection{Hypernymy as Word Classes}
\label{sec:wordnet_config}
WordNet~\cite{miller1995wordnet} is a lexical database
that groups words into sets of cognitive synonyms known as synsets,
which are in turn organized into a directed graph by various lexical relations including the hypernymy~(\emph{is-a}) relation.
As shown in Figure~\ref{fig: synset}, each vertex is a synset, labeled by the text within the box,
and each edge points from the hypernym~(supertype) to the hyponym~(subtype).
Note that a word form~(spelling) may be associated with multiple synsets -- each corresponding to a different sense of the word, which are sorted by the frequency of the sense estimated from a sense-annotated corpus.
For example, \emph{iron} has 6 synsets, among which ``iron.n.01'' is the most common one.

Hence, if two words share the same hypernym at a certain level in their hypernym-paths~(to the root in WordNet), we could say they are similar at that level.
Here we use "Depth" to quantify the hypernym-path level.
In Figure~\ref{fig: synset}, for example, at Depth 6, \emph{iron} and \emph{magnesium} are mapped to the same group named ``metallic\_element.n.01'', while \emph{desk} is mapped to ``instrumentality.n.03''.
At Depth 2, all these three words share the same (indirect) hypernym ``physical\_entity.n.01''.

In this work, we map each token in our training set into its hypernym class if this token (1) has a noun synset in the WordNet, (2) with a hypernym-path longer than a given depth $d$, and (3) has frequency below a given threshold $f$ in the training corpus.
We only consider nouns because it is not only the most common class in the WordNet but also a difficult class for LMs to learn~\cite{lazaridou2021pitfalls}.
For tokens with multiple synsets, we iterate over the synsets in the order of sense frequency and break the loop once found.
We select the most frequent synset no less than the required depth.
The mapping pseudocode is illustrated in Code~\ref{code: token2class}, which is a data pre-processing algorithm conducted only once before the training and takes no more than 5 minutes in our implementation.

\subsection{\blahfull}
We first partition the vocabulary into $\mathbf{V_{x}}$ and $\mathbf{V_{\neg x}}$ based on whether or not a token has a hypernym in the WordNet,
and $\mathbf{V_{h}}$ denotes the set of all hypernyms.
The original task in a Transformer-based LM is then to predict the token $w_j$'s probability with the output $\mathbf{x}$ from the last layer:
\begin{equation}\label{equ: org softmax}
  \begin{array}{ll}
    P(y=w_j|\mathbf{x}) = \frac{\text{exp}({\mathbf{x}^\mathsf{T}\mathbf{v}_{w_j}})}{\sum_{w_k\in\mathbf{{V_{x}}}\cup\mathbf{V_{\neg x}}} \text{exp}({\mathbf{x}^\mathsf{T}\mathbf{v}_{w_k}})}
  \end{array}
\end{equation} 
where $w_k$ is the $k_{th}$ word in the original vocabulary and $\mathbf{v}_{w_k}$ is its embedding.
Here we assume the output layer weights are tied with the input embeddings. 
We call any training step predicted with Eq.~\ref{equ: org softmax} a token prediction step.

To do the \blahfull step, we replace all tokens in $\mathbf{V_{x}}$ in a batch of training data with their corresponding hypernym classes in $\mathbf{V_{h}}$. 
After the replacement, only hypernym classes in $\mathbf{V_{h}}$ and tokens in $\mathbf{V_{\neg x}}$ can be found in that batch.
Then, the LM probability prediction becomes:
\begin{equation}\label{equ: hyper softmax}
  \begin{array}{ll}
    P(y=w_j|\mathbf{x}) = \frac{\text{exp}({\mathbf{x}^\mathsf{T}\mathbf{v}_{w_j}})}{\sum_{{w}_k\in\mathbf{{V_{h}}\cup\mathbf{V_{\neg x}}}} \text{exp}({\mathbf{x}^\mathsf{T}\mathbf{v}_{w_k}})}
  \end{array}
\end{equation} 
where $w_j$ could be either a token or a hypernym class. 
We called this batch step is a \blahfull (\blah) step.

Note that Eq.~\ref{equ: hyper softmax} is different from the multi-objective learning target, where the hypernym class would be predicted separately:
\begin{equation}\label{equ: multi-obj softmax}
  \begin{array}{ll}
    P(y=w_j|\mathbf{x}) = \frac{\text{exp}({\mathbf{x}^\mathsf{T}\mathbf{v}_{w_j}})}{\sum_{{w}_k\in\mathbf{{V_{h}}}} \text{exp}({\mathbf{x}^\mathsf{T}\mathbf{v}_{w_k}})}
  \end{array}
\end{equation} 
where $w_j$ is a hypernym class.
We will elaborate on this difference in the experiment results part.

\begin{figure}[t]
  \centering
  \includegraphics[width=0.35\textwidth,height=0.3\textwidth ]{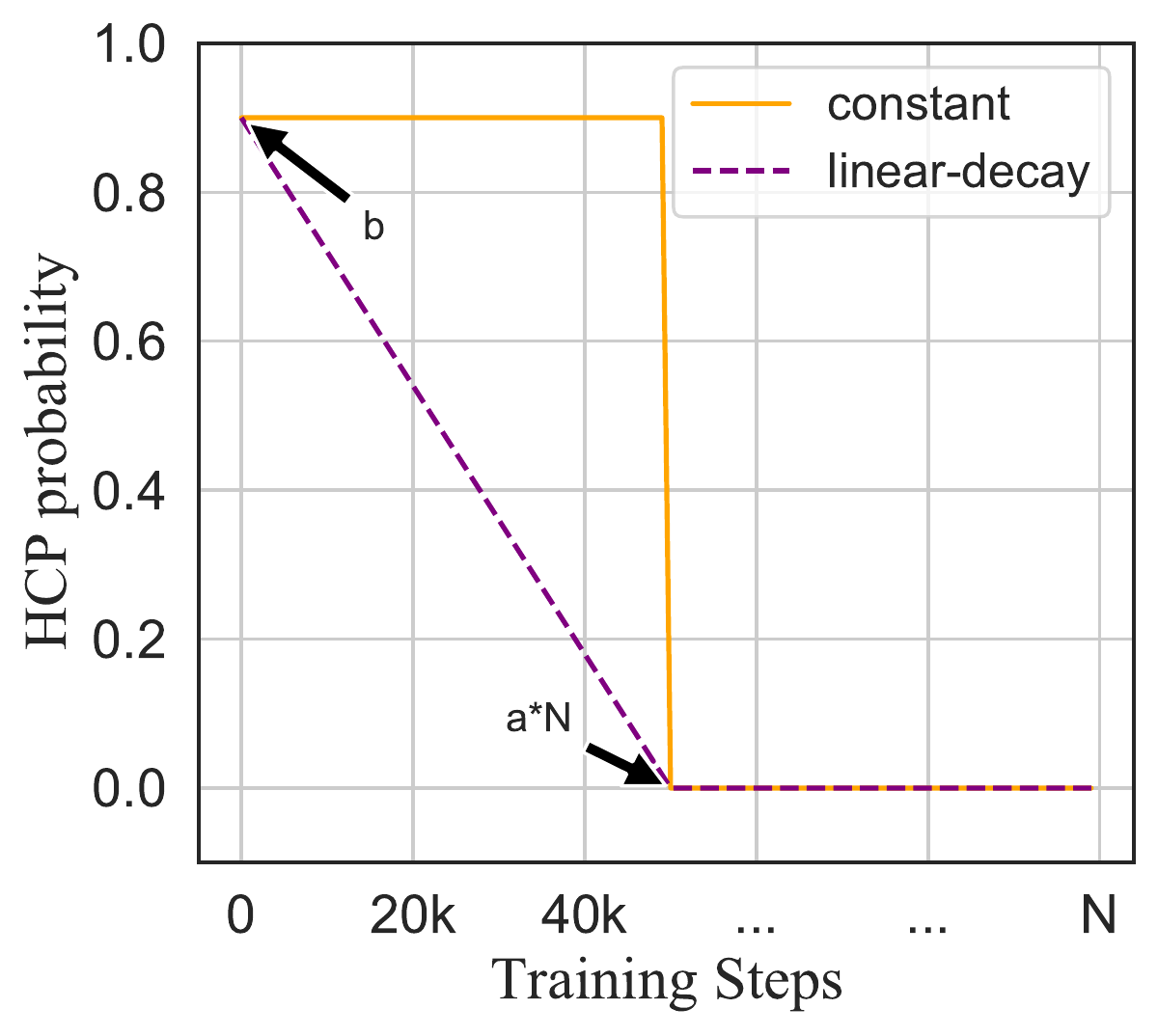}
  \caption{Probabilities of \blah step over training process with different pacing functions.}
  \label{fig: pacing function}
\end{figure}

\subsection{Training Method}
We train a LM by switching from \blah to token prediction. 
For the example in Figure~\ref{fig: synset}, our target is to teach a model to distinguish whether the next token belongs to the metallic element class or instrumentality class during the earlier stage in training, and to predict the exact word from magnesium, iron, and desk later.

Inspired by~\citet{bengio2009curriculum}, we choose curriculum learning to achieve this.
Curriculum learning usually defines a score function and a pacing function, where the score function maps from a training example to a difficulty score, while the pacing function determines the amount of the easiest/hardest examples that will be added into each epoch.
We use a simple scoring function which treats \blah as an easier task than token prediction.
Therefore, there is no need to sort all training examples.
The pacing function determines whether the current training step is a \blah step,~i.e. whether tokens will be substituted with their hypernyms.

Our pacing function can be defined as:
\begin{equation}\label{equ: constant pacing function}
  P(y=c|t) = \left\{
  \begin{array}{ll}
    b  & t<a*N \\
    0  &  t\ge a*N
  \end{array}
\right.
\end{equation} 
or
\begin{equation}\label{equ: linear pacing function}
  P(y=c|t) = \left\{
  \begin{array}{ll}
    b-b*\frac{t}{a*N}  & t<a*N \\
    0  &  t\ge a*N
  \end{array}
\right.
\end{equation} 
where $P(y=c|t)$ is the probability that the current step $t$ is a hypernym class prediction step. 
$N$ is the total training steps. 
$a$ and $b$ are hyper-parameters.
So, Eq.~\ref{equ: constant pacing function} is a constant pacing function in the first $a*N$ steps, while Eq.~\ref{equ: linear pacing function} is a linear decay function.
We plot these two functions in Figure~\ref{fig: pacing function}.
According to our experimental results Tab.~\ref{tab: pacing func }, these two functions are both effective in improving the language model. 

\begin{table*}[t]\centering
  \begin{tabular}{lccc}\toprule
  Model &\#Param. & Valid PPL&Test PPL \\\midrule
  LSTM+Neural cache~\cite{DBLP:conf/iclr/GraveJU17} &- &- &40.8 \\
  Transformer small & 91M &34.5& 36.5 \\
  \hspace{3mm} + \blah & &34.1& \textbf{35.9} \\
  \hline
  Transformer base & 151M & 29.2 & 30.7 \\
    \hspace{3mm}  + \blah &&29.1 &30.2 \\
  Transformer-XL base, M=150~\cite{DBLP:conf/acl/DaiYYCLS19} &151M&- &24.0 \\
  Segatron-XL base~\cite{Bai_Shi_Lin_Xie_Tan_Xiong_Gao_Li_2021}, M=150 &151M&- &22.5 \\
  \hspace{3mm}  + \blah &&21.9 &\textbf{22.1} \\
  \hline
  Transformer Large & 257M & 24.0 & 25.8~(80k steps) \\
    \hspace{3mm}  + \blah &&23.7 &25.3~(80k steps) \\
  Adaptive Input~\cite{baevski2018adaptive} & 247M &-&18.7~(286k steps) \\
  Transformer-XL large, M=384~\cite{DBLP:conf/acl/DaiYYCLS19} &257M &-&18.3 (400k steps) \\
  Compressive Transformer, M=1024~\cite{DBLP:conf/iclr/RaePJHL20} &257M&16.0&17.1 (400k steps) \\
  Segatron-XL large, M=384~\cite{Bai_Shi_Lin_Xie_Tan_Xiong_Gao_Li_2021} &257M &-&17.1 (350k steps) \\
    \hspace{3mm}  + \blah &&16.1 &\textbf{17.0} (350k steps) \\
  \bottomrule
  \end{tabular}
  \caption{Results on~\wikitext~dataset with different models.}
  \label{tab: main results}
\end{table*}


\section{Experiments}
We conduct experiments on two datasets. 
\wikitext~\cite{merity2016pointer} is a large word-level dataset with long-distance dependencies for language modeling.
There are 103M tokens and 28K articles (3.6K tokens per article on average).
The original vocabulary size is 271121, among which we find 3383 hypernym classes for 71567 tokens with $d=6$ and $f=6000$ (Section~\ref{sec:wordnet_config}).
\arxiv~\cite{lazaridou2021pitfalls} is collected from publicly available arXiv abstracts\footnote{https://arxiv.org/help/oa/index} with an average of 172 words per abstract and partitioned into training (1986--Sept 2017), evaluation (Aug--Dec 2017), and test (2018--2019).
Following~\citet{lazaridou2021pitfalls}, we use the BPE~\cite{sennrich2015neural} tokenization for this dataset.
The final vocabulary size is 48935, and we find 1148 hypernym classes for 5969 tokens among the vocabulary with $d=6$ and $f=1000$.

Several variants of the Transformer model have been used for our experiments:
\begin{itemize}
    \itemsep-0.2em
    \item small model: 12 layers, 10 heads, hidden size 300, batch size 256, training steps 100k;
    \item base model: 16 layers, 10 heads, hidden size 410, batch size 64,
    training steps 200k;
    \item large model: 18 layers, 16 heads, hidden size 1024 batch size 128.
\end{itemize}
The input lengths are 150 for the base model and 384 for the large model. 
The memory length is equal to the input length for both training and testing.
The hyper-parameters used for the~\arxiv~dataset are as same as the~\wikitext, except the~\arxiv~base model's input length is 384.
The number of training steps varies greatly for the large model in previous work, so we experiment on both the lower (80k) higher (350k) ends.
\begin{figure}[!t]
  \centering
\subfigure{\includegraphics[width=0.35\textwidth
]{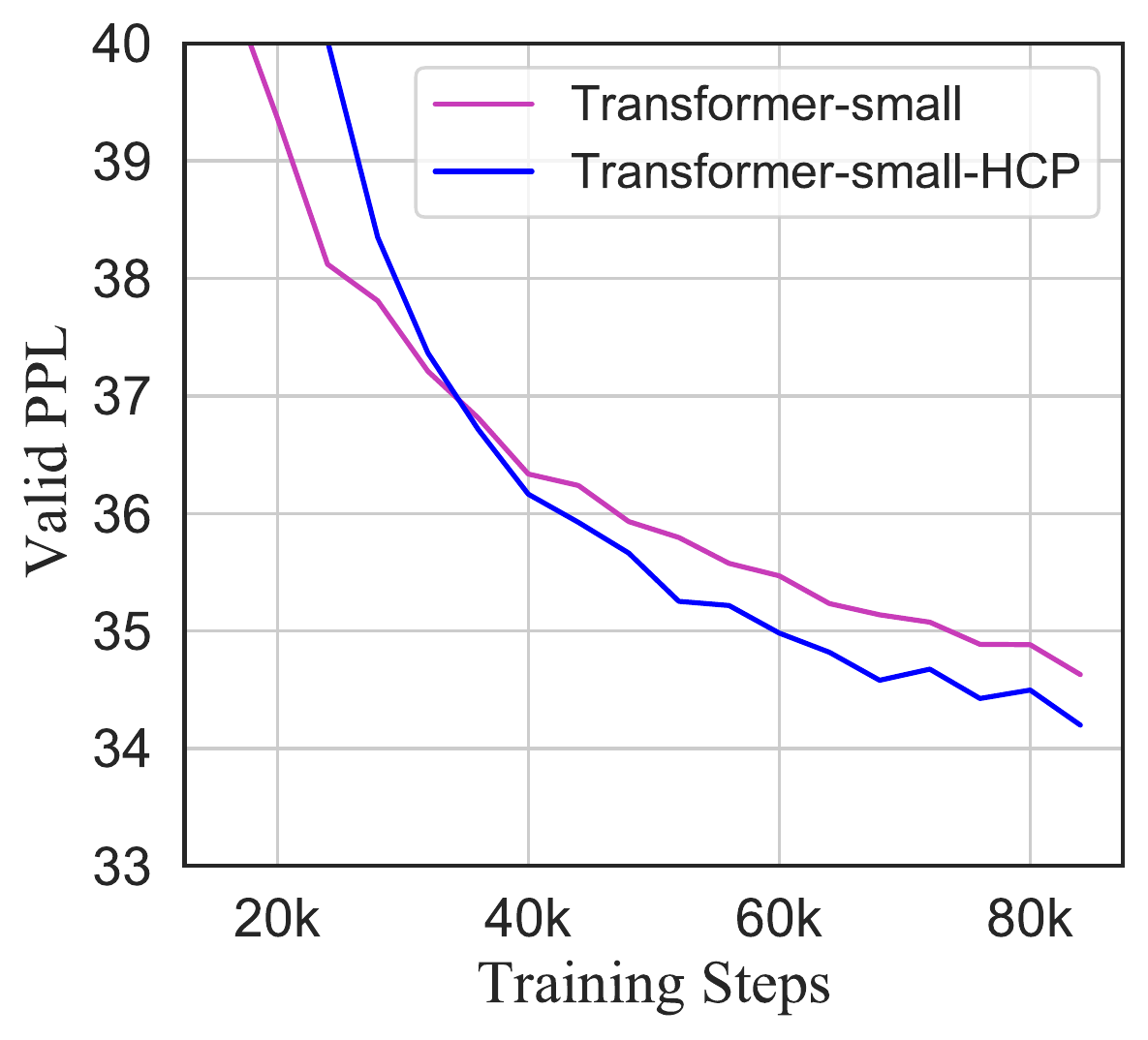}\label{fig: ppl_wiki103_small}}
\subfigure{\includegraphics[width=0.35\textwidth
]{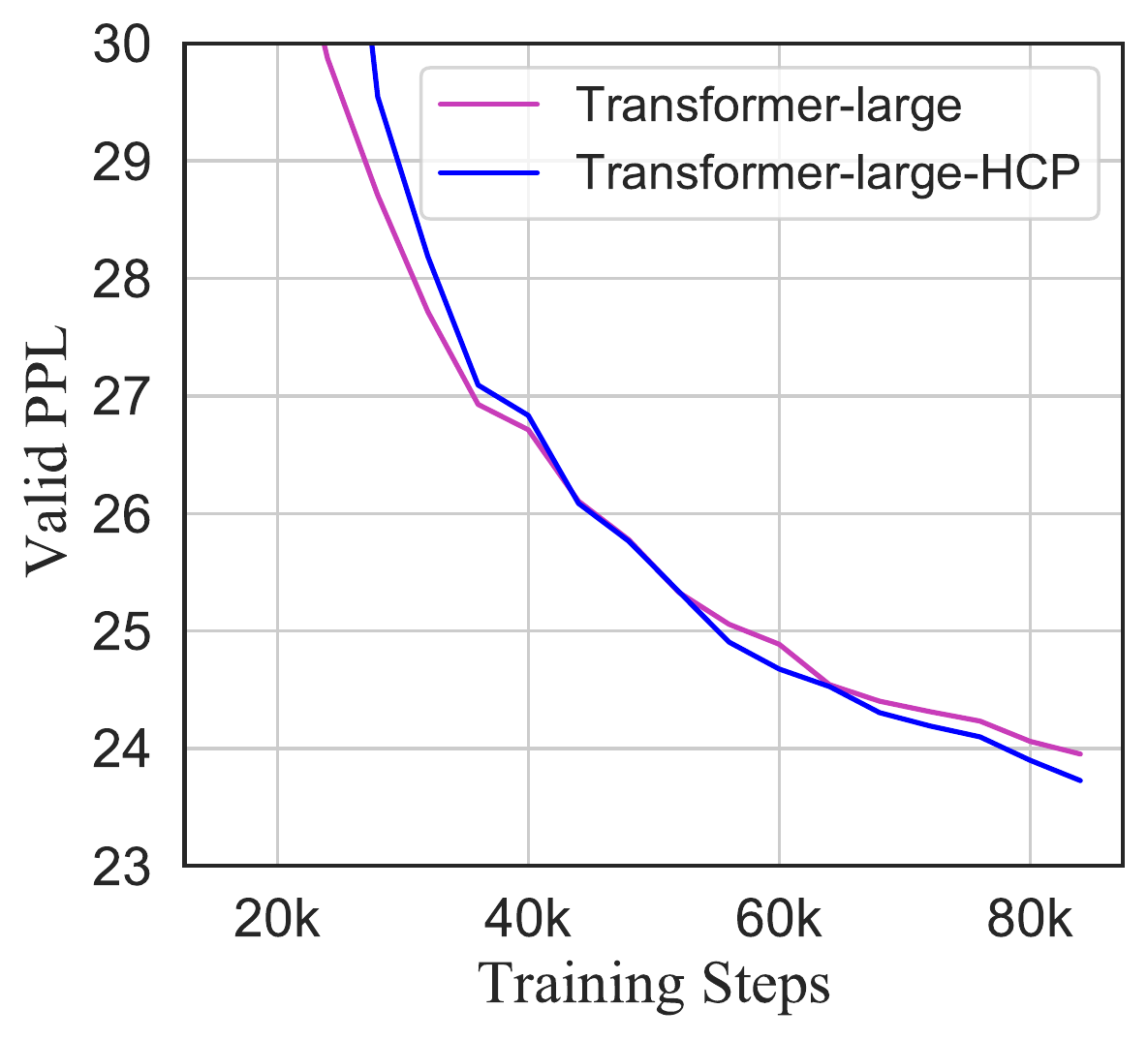}\label{fig: ppl_wiki103_large}}
  \caption{Valid perplexity curves during the training of small and large models with~\wikitext}\label{fig: ppl_wiki103}
\end{figure}

\subsection{Main results}
Our main results are shown in Table~\ref{tab: main results}.
We can see that all architectures could benefit from \blah: Transformer-small improved 0.6 ppl, Transformer-base improved 0.5, Segatron-XL base improved 0.4, Transformer-large improved 0.5, and Segatron-XL large improved 0.1. 
We also plot the validation perplexities of small and large models trained with and without \blah in Figure~\ref{fig: ppl_wiki103}.
In the beginning, the perplexity of the \blah models is higher due to the mixed training steps from the two tasks, but we can see that \blah perplexity goes down faster than the baseline method.
And after fully switching to token prediction, \blah outperforms the baseline method quickly and the gap between these two methods remains stable.
These results suggest that \blah is indeed effective in improving LM training.

\begin{table*}[!htp]
\centering
  \begin{tabular}{lccc}\toprule
  Model &\#Param. & Valid PPL&Test PPL \\\midrule
  Segatron-XL base &59M&22.39 &24.21 \\
  \hspace{3mm}  + \blah &&21.79 &\textbf{23.46} \\
  \hline
  Transformer-XL large~\cite{lazaridou2021pitfalls} &287M &-&23.07 \\
  Segatron-XL large &283M& 21.28 & 22.99 (80k steps) \\
    \hspace{3mm}  + \blah &283M&20.93 &\textbf{22.60} (80k steps) \\
  \bottomrule
  \end{tabular}
  \caption{Results on~\arxiv~dataset with different models.}
  \label{tab: arxiv results}
\end{table*}

\begin{figure*}[!ht]
  \centering
  \subfigure{\includegraphics[width=0.3\textwidth]{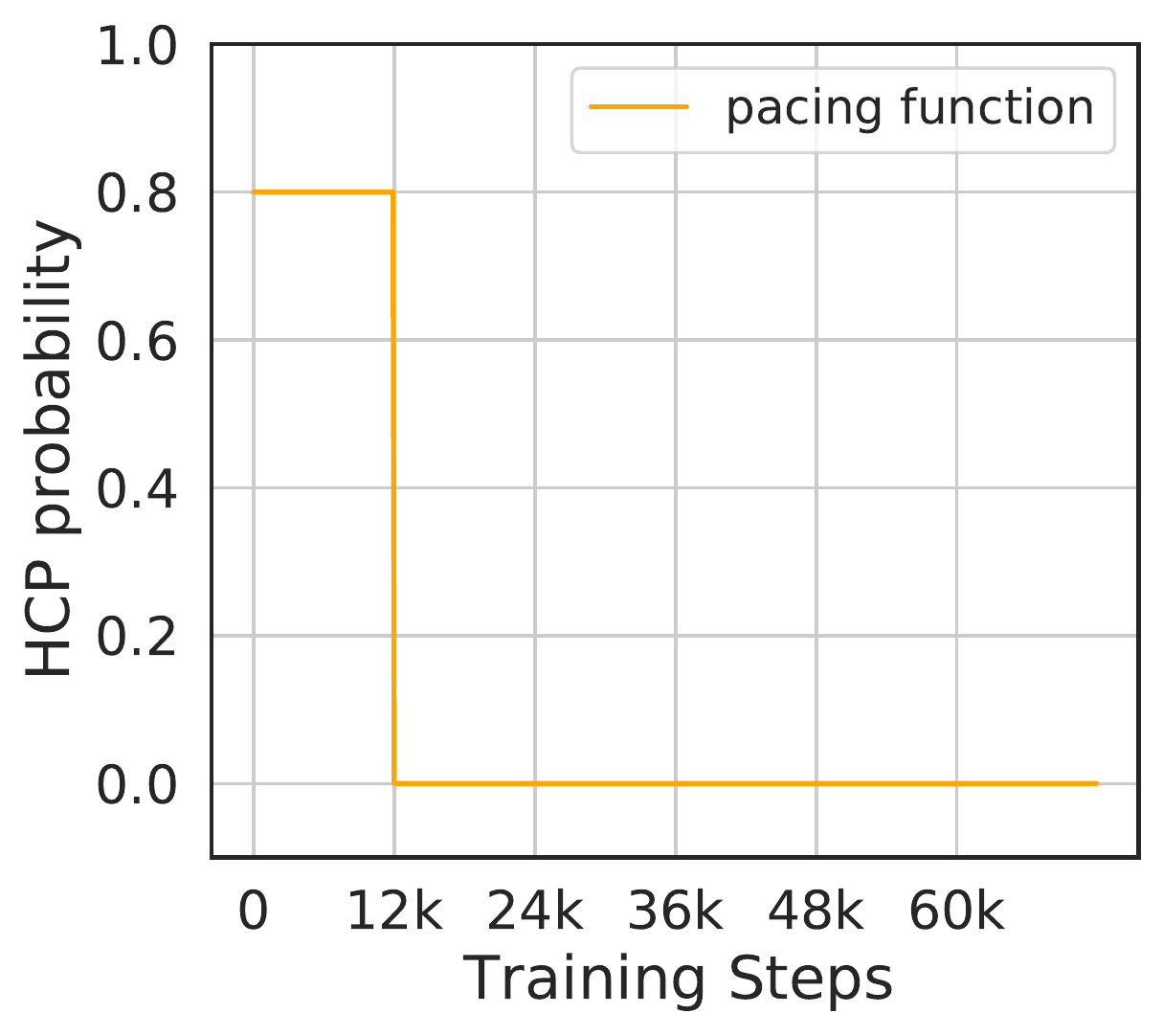}\label{fig: pacing freq}}
  \subfigure{\includegraphics[width=0.3\textwidth]{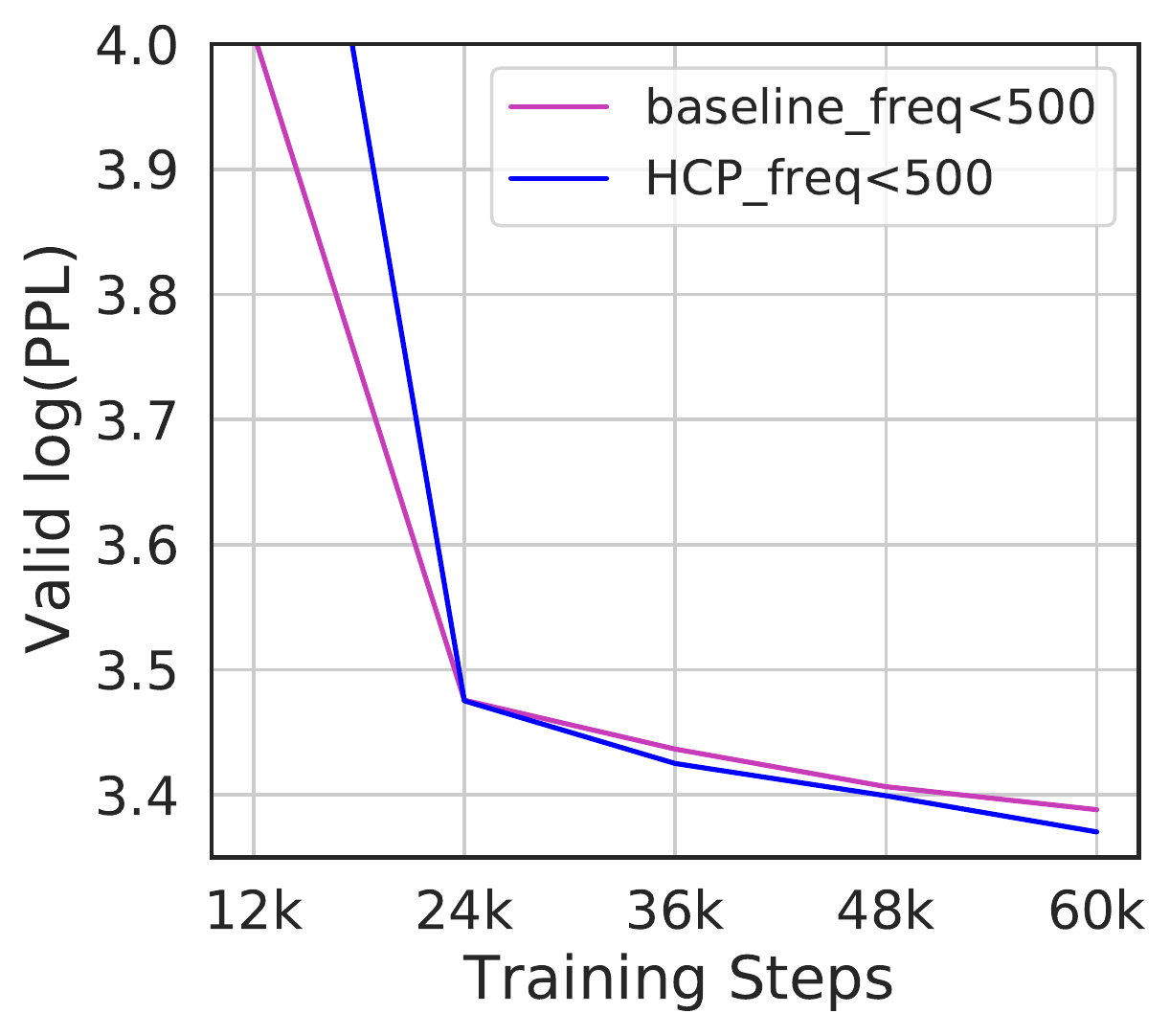}\label{fig: 500 freq}}
  \subfigure{\includegraphics[width=0.3\textwidth]{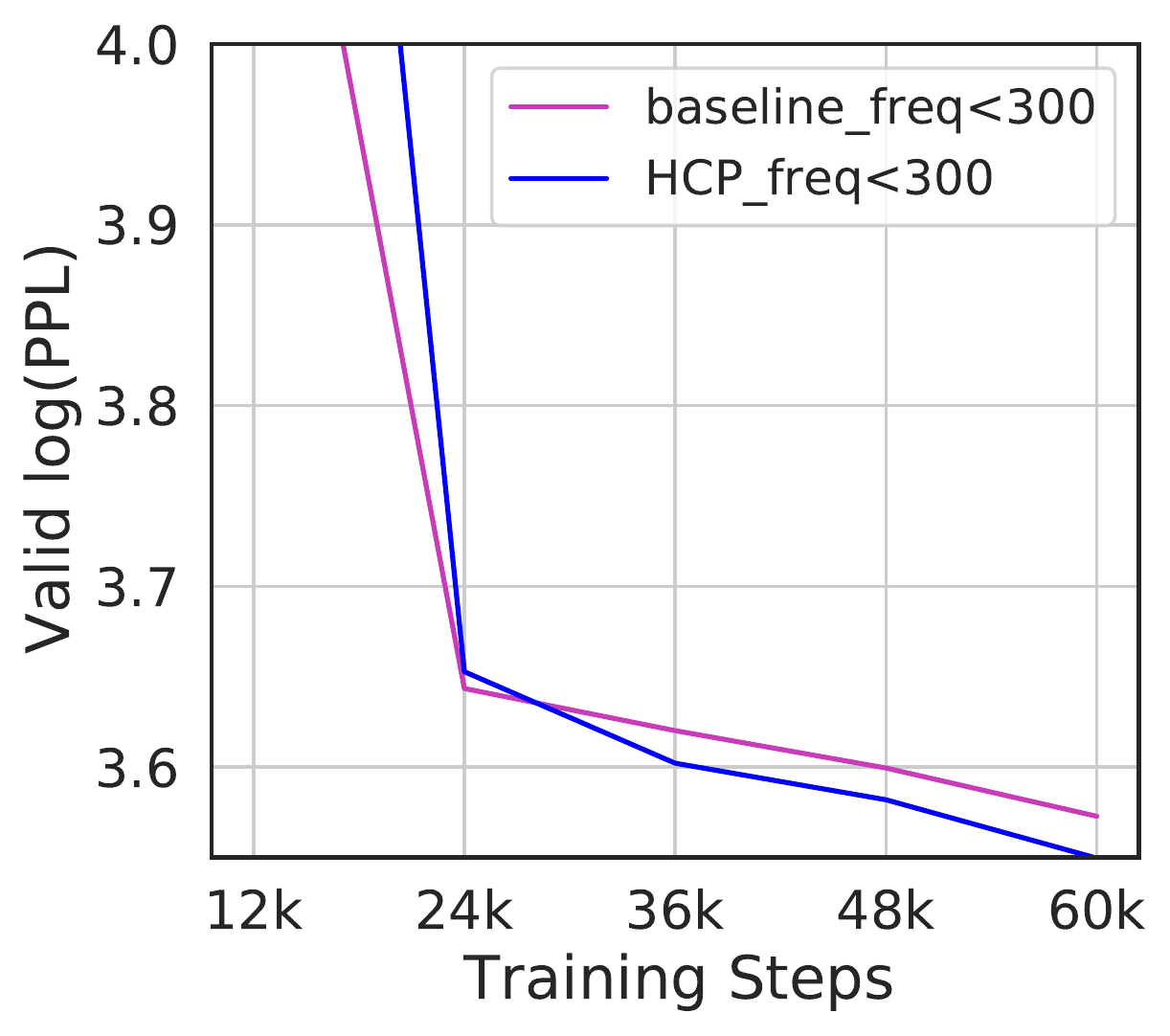}\label{fig: 300 freq}}
  \subfigure{\includegraphics[width=0.3\textwidth]{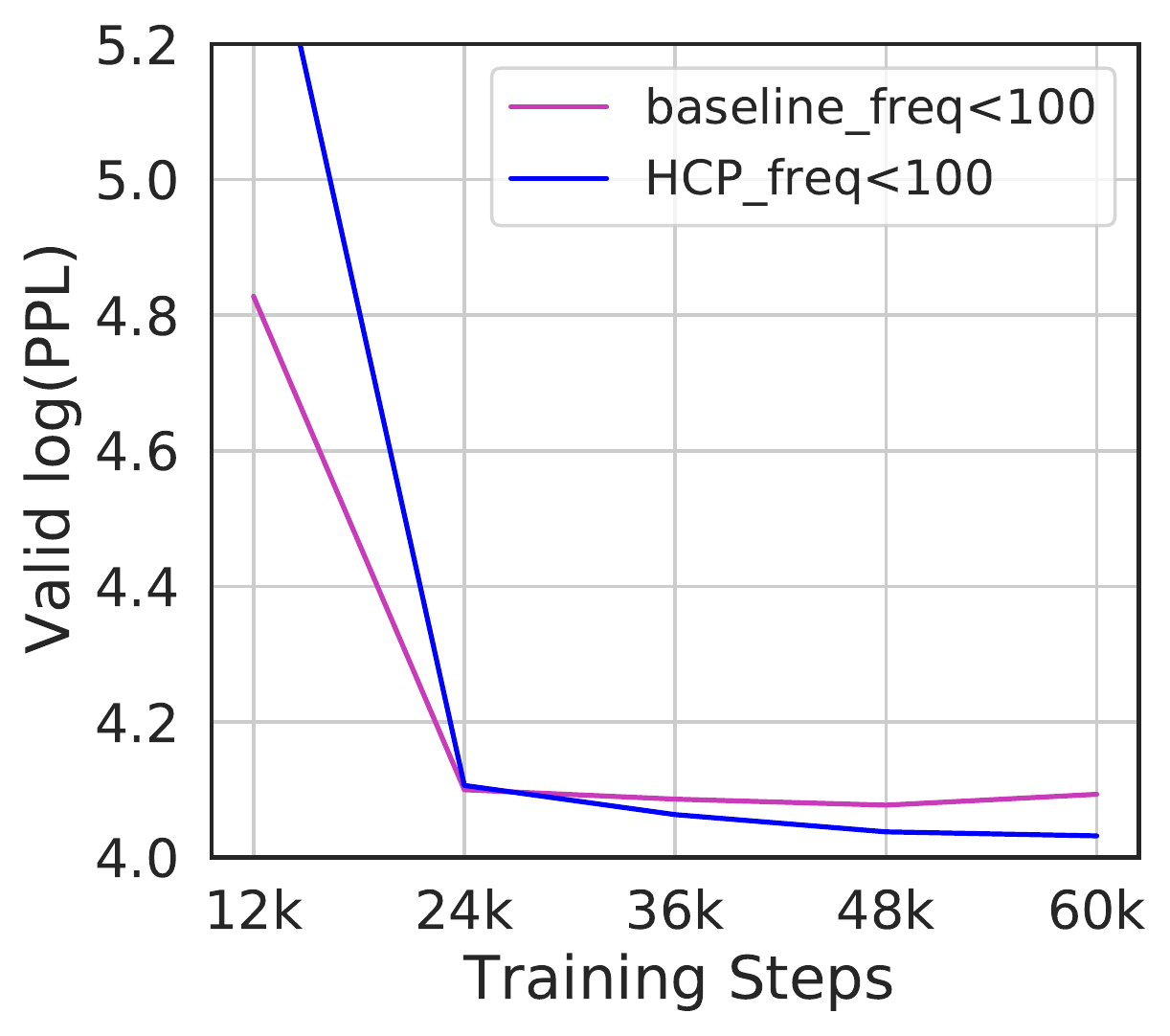}\label{fig: 100 freq}}
  \subfigure{\includegraphics[width=0.3\textwidth]{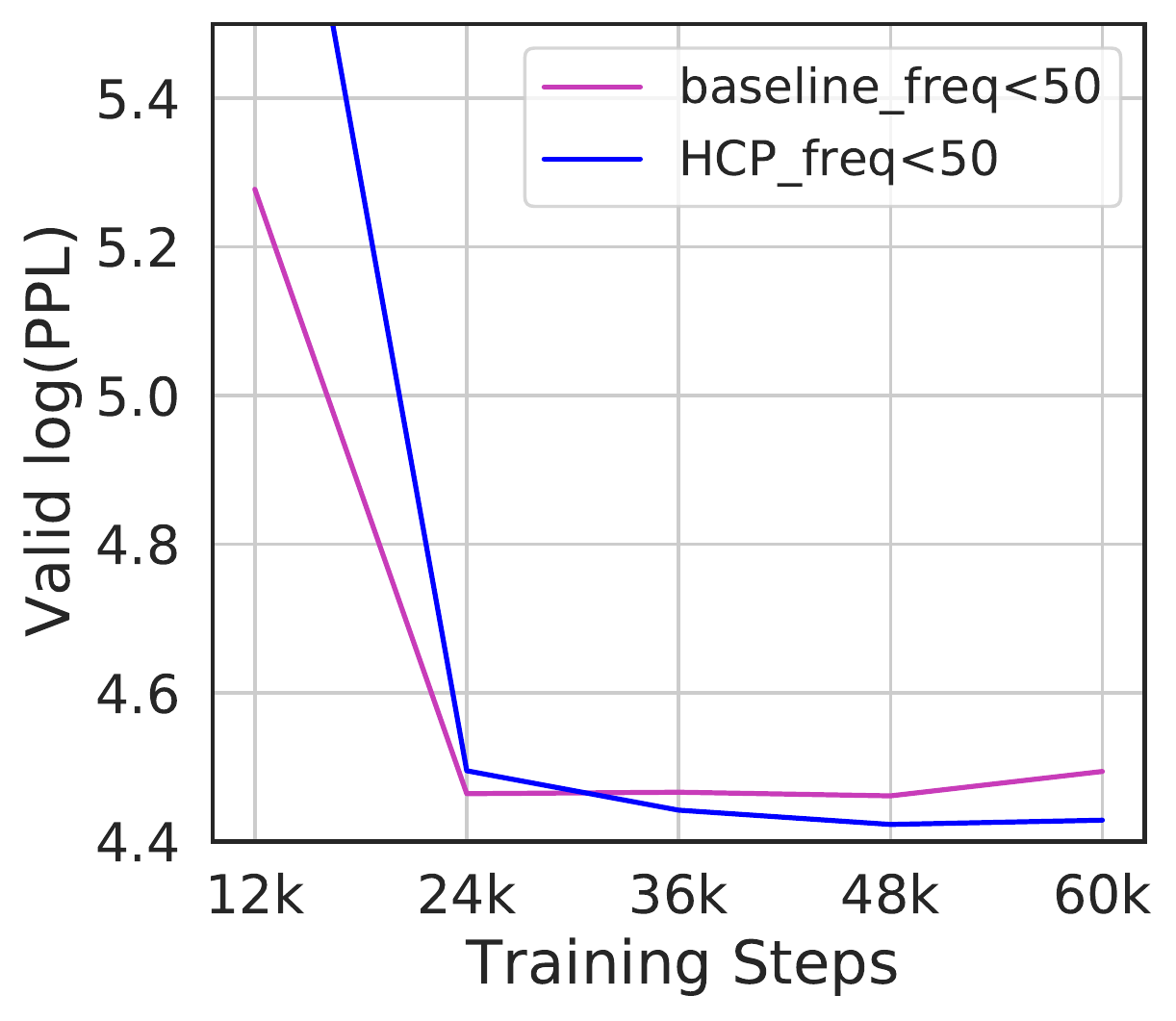}\label{fig: 50 freq}}
  \subfigure{\includegraphics[width=0.3\textwidth]{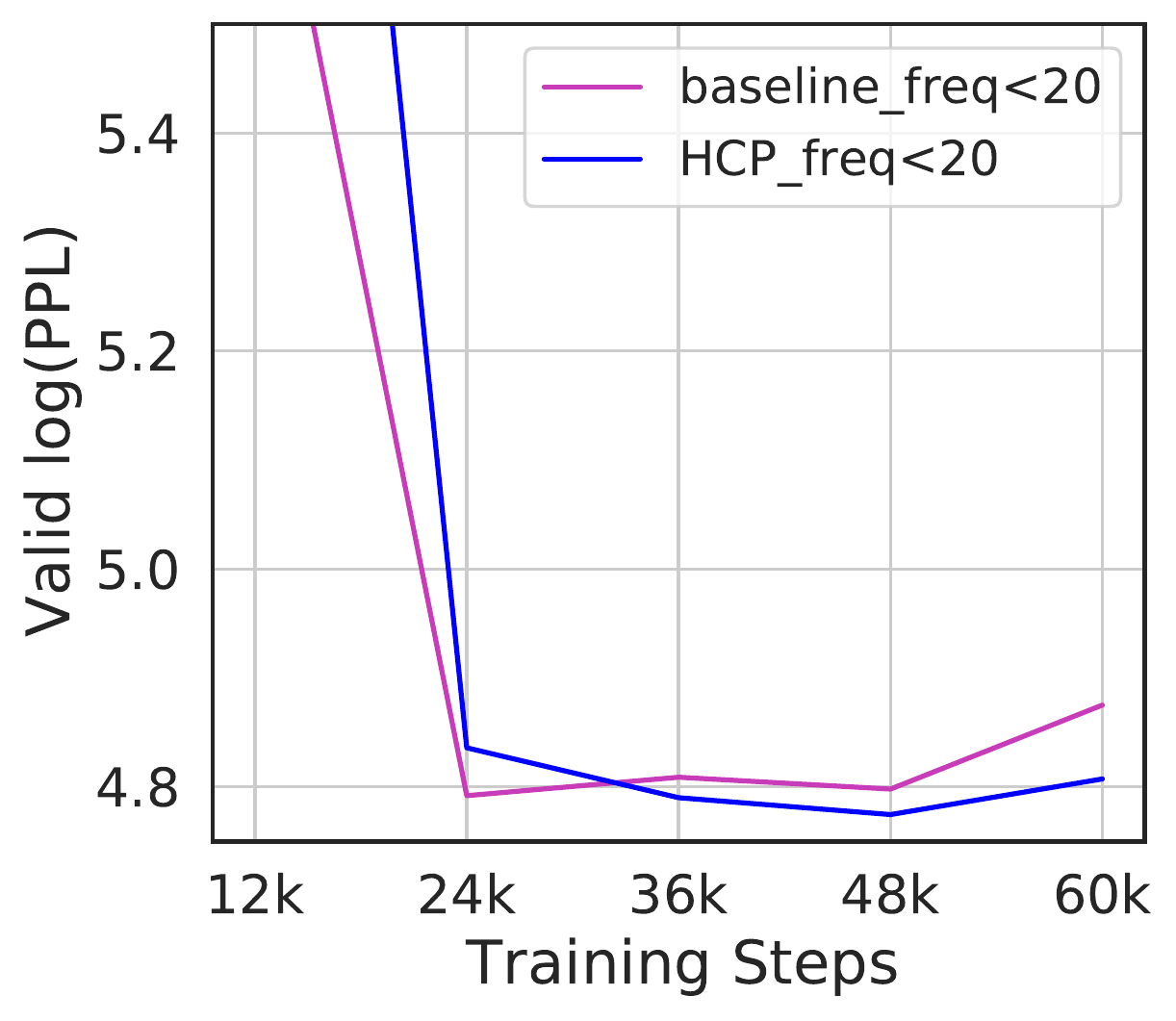}\label{fig: 20 freq}}
  \caption{Frequency-stratified validation $log(\text{perplexity})$ of baseline model~(Transformer-small) and \blah model~(Transformer-small-\blah) with \wikitext.}
  \label{fig: freq_ppl}
\end{figure*}

For experiments on the~\arxiv~dataset, we first compare the Segatron-XL base model trained with and without \blah.
The results are shown in Table~\ref{tab: arxiv results}.
The improvements over the validation set and test set are 0.6 and 0.75 respectively.
For the large model, we use the same model architecture and hyper-parameters as the~\wikitext~large model but change the vocabulary to BPE sub-tokens.
The final perplexity outperforms its counterparts about 0.4 and outperforms a larger model trained with 1024 input sequence length over 0.47, while our model length is 384.


\begin{figure*}[!ht]
  \centering
  \subfigure[Baseline model and \blah model]{\includegraphics[width=0.45\textwidth]{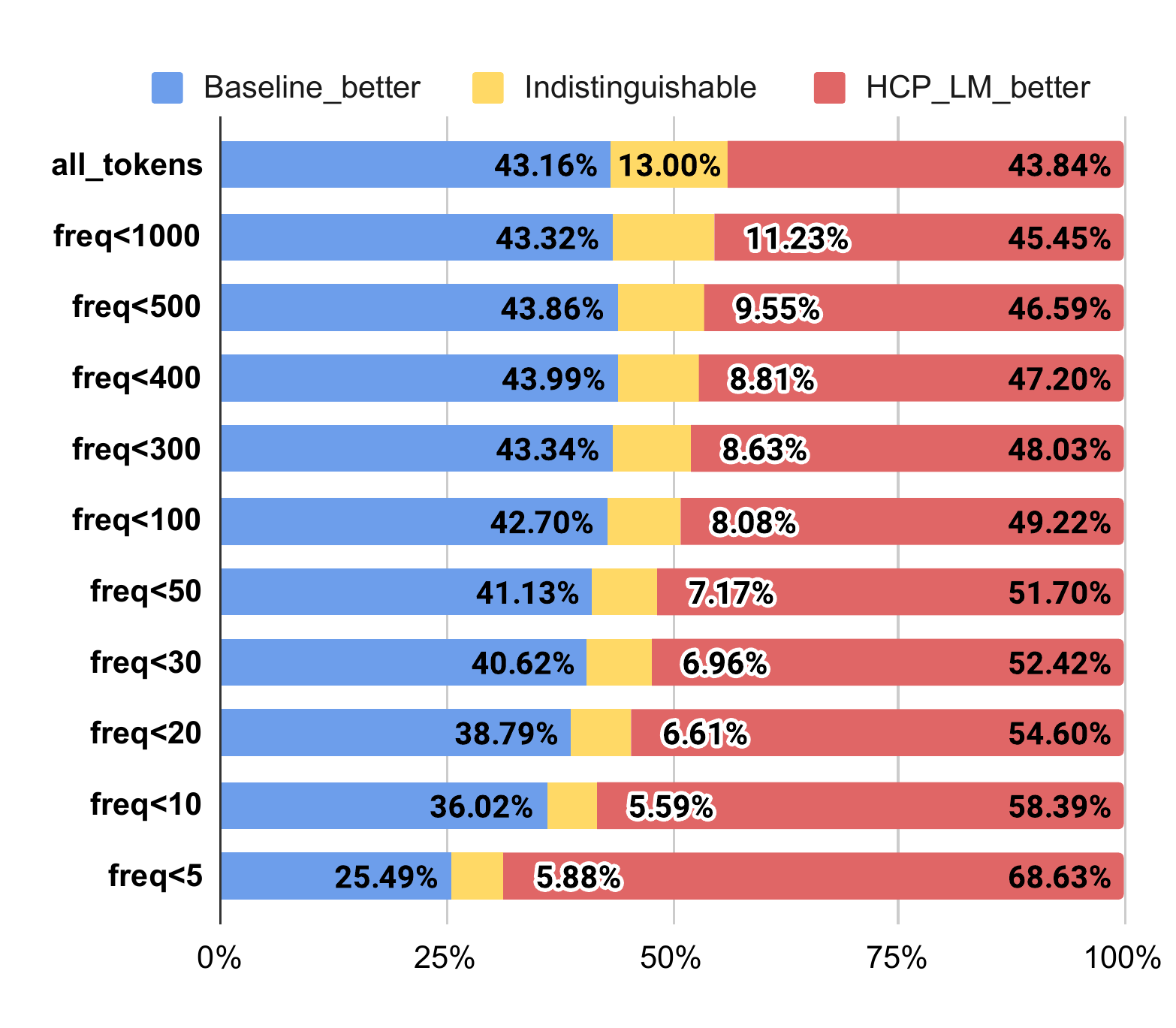}\label{fig: macro hcp}}
  \subfigure[Baseline model and sub-optimal model]{\includegraphics[width=0.45\textwidth]{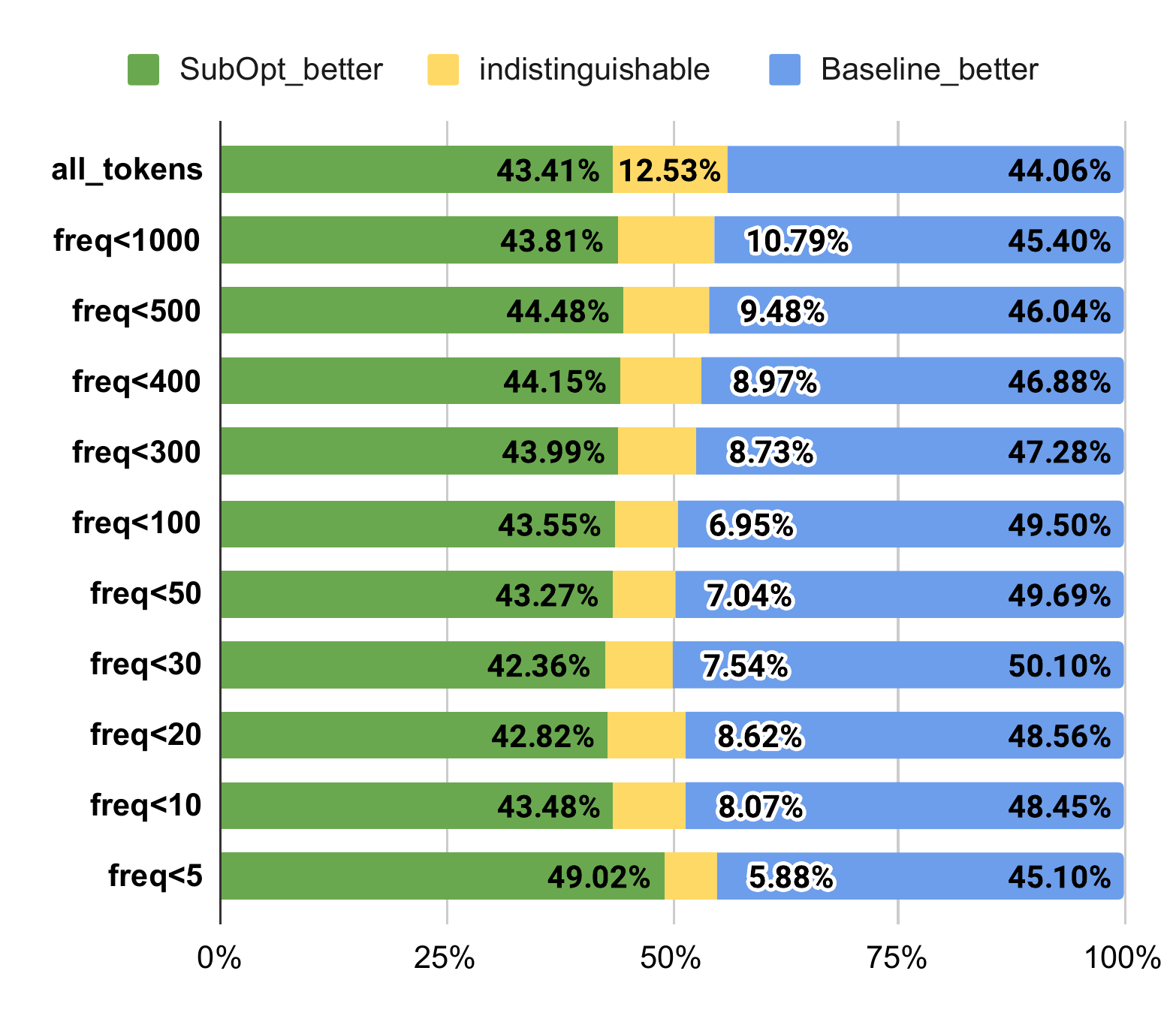}\label{fig: macro subopt}}
  \caption{Pairwise comparison results. The baseline model and \blah model are trained without and with hypernym class prediction respectively. The sub-optimal model is trained without \blah and trained with different hyper-parameters, whose perplexity is increased by 0.9 compared with the baseline model.}
  \label{fig: pairwise}
\end{figure*}

\subsection{Generalization on Rare Tokens}
In addition to the overall perplexity comparison, we also conduct comparisons with frequency-stratified validation subsets, to show the perplexity of tokens that has been replaced with the hypernym classes during training.
Results are shown in Figure~\ref{fig: freq_ppl}.
We can see that, after the first 12k hypernym class prediction steps, there is a large gap between our \blah model and the baseline model as the \blah model only learn to predict the hypernym class instead of the token itself. 
After that, in the next 12k steps, \blah's PPL decreases faster, achieves similar PPL at 24k steps, and finally outperforms the baseline method in all frequency groups.
The results show that our proposed training method can benefit the learning of the replaced tokens in various frequencies. Strikingly, we observe that, for the baseline, more training steps lead to a \emph{degradation} of performance for rare tokens, a behavior that deserves investigation in future work.

We further conduct pairwise model comparisons with tokens that have been replaced during \blah training on the \wikitext~test set. 
Given two models, we compare the prediction probabilities for each occurrence of a target token, and register a ``win'' for the model with a higher probability.
We then calculate the percentage of winnings (as well as ties) for each model by tallying over all occurrences of the token.
The results are then stratified by token frequency and plotted in Figure~\ref{fig: pairwise}.
The better model is placed on the right in both sub-figures.

In Figure~\ref{fig: macro hcp}, we see that \blah outperforms the baseline model on all frequency strata. Interestingly, \emph{the performance gap widens as frequency decreases}, indicating that \blah is beneficial in modeling rare tokens.
In Figure~\ref{fig: macro subopt}, we compare the baseline model against an under-optimized model of identical architecture but slightly different hyper-parameters.\footnote{The sub-optimal model has batch size 128 instead of the optimal 64, and the perplexity gap between these two models is observed to be slightly larger than that between \blah and the baseline (0.9 vs 0.5).}
Here, the (optimal) baseline outperforms the sub-optimal model on all but the least frequent stratum,
suggesting the possibility that perplexity reduction (resulting from hyperparameter tuning in this case) might be achieved by improving frequent word prediction \emph{at the expense of rare words}.
This is inline with observations made recently in vision tasks \citep{sagawa2020investigation}.

\subsection{Ablation study}
We conduct ablation studies with~\wikitext~dataset and Transformer small model to investigate how to map words to hypernym classes,  how to select curriculum learning pacing functions and to show why we use curriculum training.

\subsubsection{Hypernym-path Depth} 
The hypernym classes are chosen from the hypernym-paths in WordNet. 
Considering that a hypernym-path consists of multiple hypernyms, it is not straightforward to tell which layer is the best.
But the best depth $d$ should be some layer in the middle.
Because a small depth might map multiple distant words into the same class,  while a large depth will result in too many classes which are hard for a model to learn.
The extreme examples could be $d=1$ and $d=\infty$, corresponding to mapping all candidate words into the class ``Entity.n.01'' and mapping each word into itself respectively.
In Table~\ref{tab: wn layer}, we show evaluation results among different depth selections.
We find that depth 6th is the best choice, with the lowest valid perplexity.
The results also confirm our assumption that the best one would be some middle layer.
\begin{table}[!tp]
\centering
\begin{tabular}{lcc}\toprule
Depth&Valid PPL &\#Classes \\\midrule
Baseline &34.5  & 0 \\
$d=4$ &34.54  &145 \\
$d=5$ &34.29  &1169 \\
$d=6$ &\textbf{34.05}  &3383 \\
$d=7$ &34.37 &6604 \\
$d=8$ &34.25 &9063 \\
\bottomrule
\end{tabular}
\caption{Clustering words into classes with different layer's hypernym parents. The average depth is 8.03. \#Classes denotes the total number of hypernym classes.}
\label{tab: wn layer}
\end{table}

\subsubsection{Filter Frequency}
\label{sec: wn freq}
In addition to the hypernym-path depth, we also investigate how to select frequency threshold $f$.
As we mentioned above, our target is to map similar words into the same class, where predicting a hypernym class might be easier than predicting multiple different words.
After the mapping process, low-frequency words can be clustered into hypernym classes with higher frequency.
Table~\ref{tab: wn freq} shows the results of different $f$.
We can see that $f=6000$ achieves the best results while $f=\infty$~(without filter) is the worst.
We hypothesize this might be due to two reasons. First, for some high-frequency common words, the model can learn them well already, while mapping them into hypernym classes may be superfluous or even harmful. Second, including frequent words skews the marginal distribution over hypernym classes, causing hypernym prediction to be more class-imbalanced, which in turn might lead to collapsed representation in the resulting LM~\cite{Fange2103091118}. This hypothesis deserves further investigation.
It should be noted that although the difference of \#Rep.Tokens looks minor, the difference in the token's appearance is significant.
For example, $f=\infty$ maps only 776 additional tokens compared with $f=8000$, but each token's appearance is more than 8000, which explains the different perplexities in Table~\ref{tab: wn freq}.

\begin{table}[!tp]\centering
\begin{tabular}{lcc}\toprule
FilterFreq.&Valid PPL &\#Rep. \\\midrule
Baseline &34.5  & 0 \\
$f=3000$ &34.14  &70859 \\
$f=5000$ &34.50  &71735\\
$f=6000$ &\textbf{34.05}  &71971 \\
$f=7000$ &34.32 &72153\\
$f=8000$ &34.35 &72291 \\
$f=\infty$ &40.10  & 73067 \\
\bottomrule
\end{tabular}
\caption{Ignoring words whose frequency more than a threshold $f$ during hypernym class clustering. \#Rep. denotes the number of tokens in the vocabulary that will mapped.}
\label{tab: wn freq}
\end{table}

\begin{table*}[ht]
\centering
\begin{tabular}{lccccc}\toprule
Constant Func.& \blah steps &Valid PPL &NonRep.PPL &Rep.PPL \\\midrule
a=0 b=0 &0&34.5 &22.07 &348.87 \\
a=0.1 b=1 &10k&34.18&22.08 &339.30 \\
a=0.2 b=1 &20k&\textbf{34.15} &22.07 &339.34 \\
a=0.3 b=1 &30k&34.26 &22.07 &338.14 \\
a=0.4 b=1 &40k&34.39 &22.26 &338.31 \\
\midrule
Linear Func. &&  &  &  &  \\\midrule
a=0.45 b=0.45 &10k&34.14 &22.04 &340.55 \\
a=0.64 b=0.64 &20k&\textbf{34.05} &21.96 &341.33 \\
a=0.78 b=0.78 &30k&34.26 &22.05 &346.77 \\
a=0.90 b=0.90 &40k&34.56 &22.12 &354.40 \\
\bottomrule
\end{tabular}
\caption{Training N steps hypernym class prediction among 100k training steps with different pacing functions. NonRep.PPL denotes non-replaced tokens' perplexity, and Rep.PPL denotes replaced tokens' perplexity.}\label{tab: pacing func }
\end{table*}

\begin{table*}[ht]\centering
\begin{tabular}{lccccc}\toprule
&Valid PPL &Test PPL &NonRep.PPL &Rep.PPL \\\midrule
Baseline &34.50 &36.46 &22.07 &348.87 \\
Adaptive Softmax &36.32 &38.16 &22.48 &435.93 \\
Multi-obj & & & & \\
\hspace{3mm} last layer &46.06 &48.49 &27.81 &627.23 \\
\hspace{3mm} 8th layer &43.42 &45.37 &26.13 &597.66 \\
\hspace{3mm} 8th layer + mix vocab &35.97 &38.02 &22.98 &365.27 \\
\midrule
Hypernym Class Prediction &\textbf{34.05}& \textbf{35.87}&\textbf{21.96} &\textbf{341.33} \\
\bottomrule
\end{tabular}
\caption{Results obtained by alternative strategies for integrating hypernymy information into the LM: adaptive softmax and multi-objective training. Both under-perform the proposed \blah method.}\label{tab: multi-obj}
\end{table*}

\subsubsection{Pacing Function} 
Table~\ref{tab: pacing func } shows the results of models trained with various curriculum pacing functions.
We also report the validation perplexities of the tokens that have ever been replaced with hypernym class~(Rep.PPL) during training and tokens without hypernym class~(NonRep.PPL).

For the constant pacing function, we fix $b=1$ and change the value of $a$, In this case, the models are always training with \blah in the first $a*100k$ steps and then switch to the token prediction training, which is a pre-training pacing function.
We can see that all models outperform the baseline model over the validation perplexity. Rep.PPL improves from 348 to 339.
The perplexity of NonRep.PPL between baseline model and \blah models are similar, except the model trained with $a=4$, which indicates the pre-training should not take up too many steps.

For the linear pacing function, we choose some specific $a$ and $b$ to achieve the same \blah steps as the constant functions above.
For simplicity, we also set $a=b$.
In Table~\ref{tab: pacing func }, we can see that the overall perplexity of the linear functions is similar to the corresponding constant functions, where the NonRep.PPL is slightly decreased while the Rep.PPL is slightly increased.
We conduct a grid search over different pacing functions with Transformer small model and~\wikitext, and finally, use the constant function with $a=0.12$ and $b=0.8$ for all base models and large models. 

Curriculum hyper-parameters could be transferred to the~\arxiv~dataset successfully. However, we tune the frequency threshold $f$ on each dataset, because different tokenization methods change the frequency distribution. All \blah models in Table~\ref{tab: arxiv results} are using $d=6$, $f=1000$, and the constant pacing function with $a=0.12$ and $b=0.8$.

\subsubsection{Other Training Objectives}
We also experimented with two other methods to incorporate hypernym information into LM training.
Although neither method has yielded any empirical gain, we nonetheless report these methods and offer possible explanations for their failure.

\paragraph{Multi-objective Training}
Multi-objective (or multi-task) training consists in a weighted sum of token and hypernym prediction losses. We set the weight of the hypernym prediction loss to 0.2. The prediction of a token is calculated with Eq.~\ref{equ: org softmax}. The prediction of a hypernym class is calculated with Eq.~\ref{equ: multi-obj softmax}, where $\mathbf{x}$ can be the output vector from any layer in the Transformer LM.
Table~\ref{tab: multi-obj} lists the results using the last layer and the 8th layer. Using the last layer significantly undermines the original token prediction results. Using the 8th layer is better but the final perplexity is still no better than the baseline model. Simply forcing the language model to predict the hypernym class for each token is harmful to LM performance. We also tried to replace Eq.~\ref{equ: multi-obj softmax} with Eq.~\ref{equ: hyper softmax}, by mixing $\mathbf{V_{h}}$ and $\mathbf{V_{\neg w}}$ together when predicting the hypernym classes (mix vocab). This significantly improves multi-objective training.
Learning to predict the hypernym class from a mixed vocabulary $\mathbf{V_{h}}\cup\mathbf{V_{\neg w}}$ is better than only hypernym classes $\mathbf{V_{h}}$.

\paragraph{Adaptive Softmax}
Another method is the adaptive-softmax~\cite{pmlr-v70-grave17a}, where the model first predict the hypernym probability among $\mathbf{V_{h}}\cup\mathbf{V_{\neg w}}$ and then predict the token probability among the tokens with the same hypernym class.
In Table~\ref{tab: multi-obj}, we can see that the adaptive-softmax is no better than the multi-objective trained model.
By looking into the poor perplexity of Rep.PPL, we find this method cannot improve the prediction of tokens in $\mathbf{V_{w}}$.
We believe this is due to the noise of hypernym class mapping, where we choose the first synset path as the token's hypernym synset without considering the context.
Such noise will affect the adaptive-softmax prediction but is not an issue for curriculum training as the final training stage is fully trained with the original text.

\section{Conclusion}
In this work, we propose a new LM training strategy with WordNet's super-subordinate relation and curriculum learning.
Although WordNet is an external resources, it's not clear how to get lower perplexity using WordNet before this work.
Consistent perplexity reduction can be observed over various models. 
Both rare and frequent tokens can be modeling better with our proposed method while other optimization method may sacrifice the performance on rare tokens.

We'd like to address the limitations of this work: other methods to map words to classes; LM experiments with other languages; pre-training LM with our proposed method and testing on downstream tasks. 
We hope to investigate these directions in the future.

\clearpage
\bibliography{custom}
\bibliographystyle{acl_natbib}
\end{document}